%% file: ms.tex
\def\year{2020}\relax
\documentclass[letterpaper]{article} 
\usepackage{aaai20}  
\usepackage{times}  
\usepackage{helvet} 
\usepackage{courier}  
\usepackage[hyphens]{url}  
\usepackage{graphicx} 
\usepackage{graphicx}  
\usepackage{amsmath}
\usepackage{natbib}

\usepackage{hyperref}
\usepackage{pifont}

\usepackage{color}
\usepackage{float}
\usepackage{arydshln}
\usepackage{booktabs}
\usepackage{pifont}
\usepackage{xspace}
\usepackage[nameinlink,capitalize]{cleveref}
\usepackage[font=small]{caption}
\usepackage{todonotes}

\input{sub-parts/code-style.tex}

\usepackage{xcolor}
\usepackage[normalem]{ulem}
\newcommand{\gn}[1]{}

\newcommand{\eg}{\hbox{\emph{e.g.}}\xspace}
\newcommand{\ie}{\hbox{\emph{i.e.}}\xspace}

\newcommand\mr{\ensuremath{\bm{z}}\xspace}
\newcommand\x{\ensuremath{\bm{q}}\xspace}
\newcommand\env{\ensuremath{\bm{k}}\xspace}
\newcommand\ans{\ensuremath{\bm{a}}\xspace}

\newcommand{\argmax}{\operatornamewithlimits{argmax}}

\renewcommand{\tt}[1]{\fontfamily{cmtt}\selectfont #1}

\newcommand{\method}{\textsc{Wassp}\xspace}

\title{Merging Weak and Active Supervision for Semantic Parsing}
\author{ 
Ansong Ni, Pengcheng Yin, Graham Neubig \\ 
Carnegie Mellon University \\
\texttt{\{ansongn, pcyin, gneubig\}@cs.cmu.edu}
}
 
\begin{document}

\maketitle

\input{sub-parts/1-abstract-intro}

\input{sub-parts/2-approach-overview.tex}

\section{Query Sample Selection Heuristics}
\input{sub-parts/3-sample-selection.tex}

\section{Forms of Extra Supervision}
\input{sub-parts/4-extra-supervision.tex}

\section{Experiments}
\input{sub-parts/5-experiments.tex}

\section{Related Work}
\input{sub-parts/related-work.tex}

\section{Conclusion}
We propose \method, a framework to merge weak and active learning for semantic parsing. We study different query sample selection heuristics and various forms for extra supervision. Experiments on two datasets show that \method can greatly improve the performance of a weakly-supervised semantic parser with a small fraction of examples queried.

\section{Acknowledgements}
The authors thank Eduard Hovy, Alessandro Oltramari, and anonymous reviewers for their advice regarding this research \gn{Please add any others that you know of.}.
This work was supported in part by a gift from the Carnegie Bosch Institute.

\bibliography{ms}
\bibliographystyle{aaai}

\end{document}

%% file: sub-parts/code-style.tex
\usepackage{booktabs}
\usepackage{graphicx}
\usepackage{float}
\usepackage{color}
\usepackage{amsmath,amsfonts,amssymb,bbm,epsfig,bm}
\usepackage[ruled,noresetcount,linesnumbered]{algorithm2e}
\usepackage{listings}
\usepackage{subfig}
\usepackage{url}
\usepackage{spverbatim}
\usepackage{multirow}
\usepackage{pbox}
\usepackage{tikz}
\usepackage{pifont}
\usepackage{xspace}
\usepackage{subfig}
\usepackage{physics}
\usepackage[nameinlink,capitalize]{cleveref}
\usepackage{scalerel}
\usepackage[font=small]{caption}
\usepackage{titlesec}
\usepackage{todonotes}
\usepackage[normalem]{ulem}

\usepackage{color}
\definecolor{deepblue}{rgb}{0,0,0.5}
\definecolor{deepred}{rgb}{0.6,0,0}
\definecolor{deepgreen}{rgb}{0,0.5,0}
\definecolor{darkgreen}{RGB}{43,163,39}
\definecolor{bluesquare}{rgb}{126,166,224}

\newcommand{\cmark}{\textcolor{darkgreen}{\ding{51}}}
\newcommand{\xmark}{\textcolor{red}{\ding{55}}}

\lstdefinestyle{pythoncode}{
  language=Python,
  morekeywords={self,join,append,split,write,chain},             
  keywordstyle=\bfseries\color{deepblue},
  emph={MyClass,__init__,assert,return,raise},          
  emphstyle=\color{deepred},    
  showstringspaces=false,
  breaklines=true,
  escapeinside=||,
  columns=fullflexible,
  basicstyle=\fontfamily{cmtt}\small,
  belowskip=0.2\baselineskip,
  aboveskip=-0.6\baselineskip
}

\lstdefinestyle{atiscode}{
  language=lisp,
  deletekeywords={top},
  morekeywords={self,argmax,argmin,flight,from,to,departure,time,exists,round,trip,fare,has_meal,during,day,day,number,month,ground,transport,to,city,from,airport,meal,weekday,oneway,round_trip,day,rental,car,ground,fare,has,meal,nonstop,round,round_trip},
  keywordstyle=\bfseries\color{deepblue},
  emph={lambda},          
  emphstyle=\color{deepred},    
  showstringspaces=false,
  breaklines=true,
  escapeinside=||,
  columns=fullflexible,
  basicstyle=\fontfamily{cmtt}\small,
  belowskip=0.2\baselineskip,
  aboveskip=-0.5\baselineskip
}

\lstdefinestyle{nsmcode}{
  language=lisp,
  deletekeywords={top},
  morekeywords={all_rows, v0},
  keywordstyle=\bfseries\color{deepblue},
  emph={filter_eq, hop, count, filter_less, filter_greater, argmax},          
  emphstyle=\color{deepred},    
  showstringspaces=false,
  breaklines=true,
  escapeinside=||,
  columns=fullflexible,
  basicstyle=\fontfamily{cmtt}\small,
  belowskip=0.2\baselineskip,
  aboveskip=-0.5\baselineskip
}

\renewcommand{\tt}[1]{\fontfamily{cmtt}\selectfont #1}

%% file: sub-parts/1-abstract-intro.tex
\begin{abstract}
A semantic parser maps natural language commands (NLs) from the users to executable meaning representations (MRs), which are later executed in certain environment to obtain user-desired results. The fully-supervised training of such parser requires NL/MR pairs, annotated by domain experts, which makes them expensive to collect. However, weakly-supervised semantic parsers are learnt only from pairs of NL and expected execution results, leaving the MRs latent. While weak supervision is cheaper to acquire, learning from this input poses difficulties. It demands that parsers search a large space with a very weak learning signal and it is hard to avoid spurious MRs that achieve the correct answer in the wrong way. These factors lead to a performance gap between parsers trained in weakly- and fully-supervised setting. To bridge this gap, we examine the intersection between weak supervision and active learning, which allows the learner to actively select examples and query for manual annotations as extra supervision to improve the model trained under weak supervision. We study different active learning heuristics for selecting examples to query, and various forms of extra supervision for such queries. We evaluate the effectiveness of our method on two different datasets. 
Experiments on the WikiSQL show that by annotating only 1.8\% of examples, we improve over a state-of-the-art weakly-supervised baseline by 6.4\%, achieving an accuracy of 79.0\%, which is only 1.3\% away from the model trained with full supervision. Experiments on WikiTableQuestions with human annotators show that our method can improve the performance with only 100 active queries, especially for weakly-supervised parsers learnt from a cold start.%
\footnote{Code available at \href{{https://github.com/niansong1996/wassp}}{https://github.com/niansong1996/wassp}}
\end{abstract}

\section{Introduction}
Semantic parsing maps a user-issued natural language (NL) utterance to a machine-executable meaning representation (MR), such as $\lambda-$calculus \citep{zettlemoyer2005learning}, SQL queries \citep{dong2018coarse} or general purpose programming languages (\eg, Python) \citep{yin2018tranx}. 
These MRs can then be executed in a certain environment (\eg, a knowledge base, KB) to achieve the goal of users, such as querying a KB using natural language. 

Classical supervised learning of semantic parsers requires parallel corpora of NL utterances and their corresponding annotated MRs \citep{zettlemoyer2005learning,dong2018coarse,rabinovich2017abstract}.
However, this annotation requires strong domain expertise in the schema of the MR language (
\eg, SQL of KB queries), and hiring such domain experts can be expensive.
Moreover, depending on the type of meaning representation used, the same input utterance could be grounded to multiple, equally acceptable MRs with diverse syntactic form (\eg different ways to implement a loop), making it non-trivial to validate the correctness of annotated MRs given by different annotators.
These factors make it very expensive to create fully-supervised semantic parsing datasets at large scale. 

\begin{figure*}
    \centering
    \includegraphics[width=17.7cm]{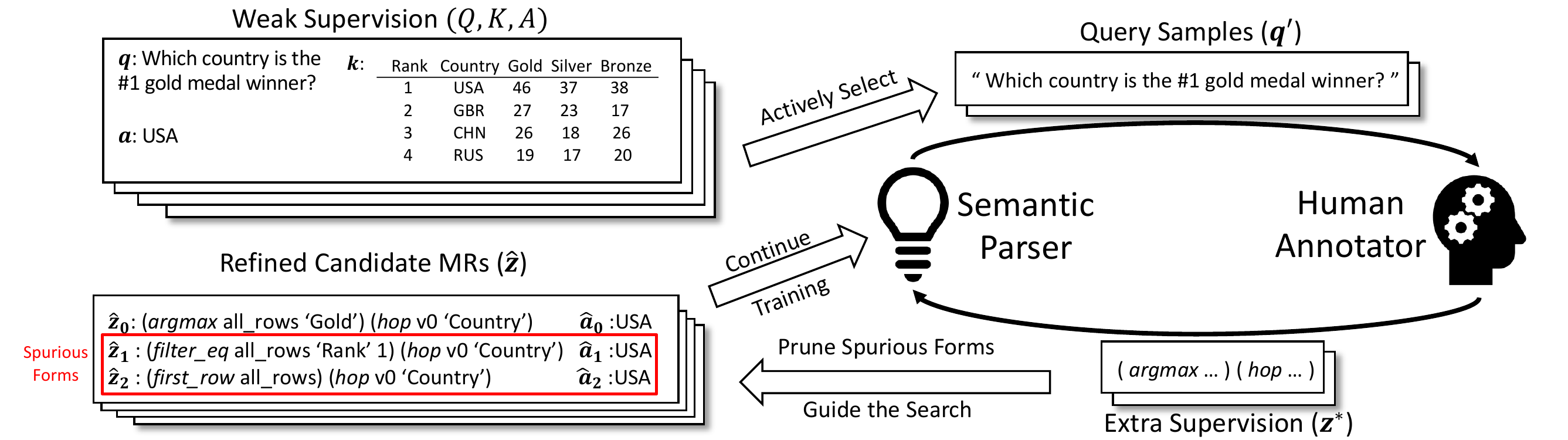}
    \caption{An overview of our proposed \method: we first train a semantic parser with weak supervision till convergence, then a small fraction of NLs are selected from the training set to actively query the annotator for extra supervision, which is then used to refined the candidate MRs for the next round of weakly-supervised training.}
    \label{fig:wassp_overview}
\end{figure*}

\emph{Weakly-supervised} semantic parsing aims to solve this problem by relaxing the requirement for annotated MRs, instead training the parser using indirect supervision of \emph{only} the expected result of executing an MR corresponding to user's intent (\ie, an answer for a user question) \citep{clarke2010driving,berant2013freebase,pasupat2015compositional}.
Compared with annotation-intensive MRs, such expected results are easy to obtain, even from annotators who do not know the specific MR language.
Still, weakly supervised learning of semantic parsers remains a non-trivial task.
First, the search space of possible MRs given an execution result is exponentially large, while the reward the parser receives during training is sparse and binary (\eg one \textit{iff} the execution result is correct);
Another issue is \emph{spurious MRs}, which are programs that happens to execute to the correct result, but are semantically incorrect.  
For example, a parser may wrongly parse the utterance ``multiply two by two" to ``2+2" and still get the correct answer.
Such spuriousness would add a great amount of noise to the already-weak supervision signal.

To tackle these challenges with weakly-supervised semantic parsing, we propose a method to combine \underline{W}eak and \underline{A}ctive \underline{S}upervision for \underline{S}emantic \underline{P}arsing (\method). 
As illustrated in Fig.~\ref{fig:wassp_overview}, for each iteration, we first train the semantic parsing model to convergence with weak supervision. 
Then we perform active learning which allows the model to actively ask for extra supervision (\eg annotated MRs) for only the small fraction of examples that would maximally improve its performance.
Finally, we feed the extra supervision to the semantic parsing model to rule out spurious logical forms and guide the search for hard examples. 

Given our proposed \method framework, we examine three main research questions:
\begin{enumerate}
    \item[RQ1] How effective is \method in improving the performance of weakly-supervised semantic parser?
    \item[RQ2] How can we pick the set of utterances where direct supervision will maximally benefit parsing accuracy?
    \item[RQ3] What kind of extra supervision can \method use to improve the model for each sampled utterance? 
\end{enumerate}
Specifically, to answer RQ1, we measure the improvement of \method over a weakly-supervised baseline with varying number of queries made.
To answer RQ2, we investigate different heuristics to select a smallest set of NL utterances for annotation, by considering the \textit{informativeness}, \textit{representativeness} and \textit{diversity} of an example.
To answer RQ3, we study various forms of extra supervision to provide to the model for each sampled utterance.
While just presenting the corresponding fully-specified MRs would be a trivial strategy, we show that this is not our only choice.
Instead, using a simpler form, the \textit{sketch} of an MR, the actively-supervised semantic parsers can achieve similar performance compared to using the resource-heavy MRs. 


We evaluate the effectiveness of our proposed active learning approach on two different datasets: the WikiSQL dataset \citep{zhong2017seq2sql} and the WikiTableQuestions dataset \citep{pasupat2015compositional}.
We demonstrate that active learning can greatly improve the performance semantic parsers trained with weak supervision with only a small fraction of examples actively selected and queried: By querying a total fraction of 1.8\% of the training examples for WikiSQL, we improve over a state-of-the-art weakly-supervised baseline by 6.4\%, achieving an accuracy of 79.0\%, which is only 1.3\% away from the model trained with full supervision. Experiment results on WikiTableQuestions show that \method can boost the performance of a model suffered from a cold start by 25.6\% with only 50 examples. 
We also show that alternative forms of extra supervision to fully-specified MRs as MR sketches may also be used by \method and yields similar performances, only lose an maximum of 1.4\% of accuracy comparing to full gold MR annotations on the WikiSQL dataset.%


%% file: sub-parts/2-approach-overview.tex
\section{Weakly Supervised Semantic Parsing}
\subsection{Formulation}
Formally, given a natural language utterance \x, a semantic parser is a parametric model $P(\mr|\x ;\theta)$ that transduces the utterance \x into a meaning representation \mr. The parser could be optimized using either classical supervised learning, or weakly supervised learning.

In the classical setting of (fully-) supervised learning, a model is optimized using a parallel corpus $D$ consisting of paired NL utterances and their annotated MRs, \ie, $D=\{
\langle \x_i, \mr_i \rangle \}_{i=1}^{N}$. The standard maximum likelihood learning problem is:
\begin{equation}
\label{fssp}
    \argmax_{\theta} \prod_{\langle \x_i, \mr_i \rangle \in D} P(\mr_i|\x_i;\theta).
\end{equation}

Weakly-supervised learning can be formulated as a reinforcement learning problem, instead of access to gold-standard MRs, the semantic parser (\ie, the agent) is only presented with an execution model \env (\ie, the environment, such as a database) and the gold execution result \ans. The parser interacts with \env, searching for high-reward MRs $\hat{\mr}$ that execute to the correct results (\ie, $\env(\hat{\mr}) = \ans$). 
A training instance is therefore defined by a triplet: utterance \x, execution environment \env and gold answer \ans. The learning objective is to maximize the probability of  the correct answer \ans by marginalizing over all candidate MRs that execute to \ans: 
\begin{align}
\label{wssp}
    & \argmax_{\theta} \prod_{\langle \x_i, \mr_i \rangle \in D} P(\ans_i|\x_i;\theta) \nonumber \\ 
    =& \argmax_{\theta} \prod_{\langle \x_i, \mr_i \rangle \in D} \sum_{\{\hat{\mr}_{i}\in \hat{Z} | \env(\hat{\mr}_i) = \ans_i\}} P(\hat{\mr}_i|\x_i;\theta)  
\end{align}
Weakly supervised parsers are usually trained with an EM-like method: the model searches for high-reward MRs by sampling from the semantic parser, then the model parameters are optimized to maximize the probability of generating these high-reward MRs. 

\subsection{Difficulties with Weak Supervision}
\label{section:problems}


Two challenges make weakly-supervised parsing harder than its fully-supervised counterpart:

\noindent \textbf{Exploration of an Exponentially-large Space.} 
The search space of latent MRs is exponential or infinite, making it intractable to exhaustively search over the set of plausible MRs $\{\hat{\mr}_i | \env(\hat{\mr}_i)=\ans_i\}$ to calculate sum in Eq.~\ref{wssp}.
Thus, it is common to resort to sampling or $k$-best search to approximate this space \citep{guu2017language}, potentially combined with sophisticated methods to reduce the space of plausible MRs with type constraints \citep{krishnamurthy2017neural} or use of a memory buffer to cache high-reward explored MRs \citep{liang2018memory}.
However, inferring MRs for complex, highly compositional input utterances still remains an open challenge~\citep{pasupat2016inferring}.

\noindent \textbf{Spurious MRs. } 
In contrast with the full-supervised setting where the parser is optimized using the semantically correct gold-standard MR for each utterance, the search space of latent MRs contains spurious samples that happen to execute to the correct results despite being semantically incorrect.
For example, in \autoref{fig:wassp_overview}, the NL utterance \textit{``Which country is the \#1 gold medal winner?''} has one semantically correct MR $z_0$ while MRs $z_1$ and $z_2$ do not match the semantics of the original utterance. 
Nonetheless, $z_1$ and $z_2$ follow the grammatical rules, satisfy the type constraints, and get correct results, and thus there is no simple way for a weakly supervised learner to distinguish them from gold-standard MRs.
Such spurious MRs could be exponentially many \citep{pasupat2016inferring}, and add significant noise to the training process. 
While attempts have been made to mitigate the issue of spuriousness, such as using a ranker trained with extra semantic or lexicon information \citep{cheng2018weakly} or introduce prior knowledge to bias the policy \citep{misra2018policy}, it is still highly non-trivial to totally solve this issue due to the ambiguous nature of the reward.

Empirically, these challenges lead to performance gap between fully- and weakly-supervised semantic parsers.
For example, on the WikiSQL dataset, the state-of-the-art fully-supervised semantic parser reaches an testing execution accuracy of 86.2\% \citep{hwang2019comprehensive} while the best weakly-supervised semantic parser \citep{liang2018memory} only has an accuracy of 72.6\%, a difference of 13.6\%.

\section{Merging Weak and Active Supervision}
\label{section:imp-al}



In order to solve these two problems, we first discuss three key insights about weakly-supervised semantic parsing that motivates \method:

\noindent 
1) Though mixed with spurious MRs, weakly-supervised semantic parsers are still able to find gold MRs for simpler input utterances, even with such noisy and weak learning signal. 
This means that effective training of semantic parsers does not require full supervision using annotated MRs for \emph{all} examples.

\noindent
2) The EM-like weakly-supervised optimization will ``stall'' at the point when no new high-reward MRs can be found by the parser, and the set of discovered MRs with correct answers are unchanged for all examples in the training set.
Eq.~\ref{wssp} will then converge to a local optimum. 
However, if the parser can discover new MRs with correct execution results, growing the set of high-reward MRs for even a small fraction of training examples, the optimization of Eq.~\ref{wssp} will resume, and the parser could in turn explore more similar high-reward MRs with its updated parameters.
In other words, it is possible that some extra supervision for only a small fraction of the training data can prevent stagnation and resume the optimization process.

\noindent
3) Fully-specified MRs might not always be necessary to provide extra supervision for a weakly-supervised parser to discover high-reward MRs for complex utterances or rule out spurious MRs for ambiguous inputs.
As an example, to identify the semantically correct MR for the illustrative example in \autoref{fig:wassp_overview}, the parser only need to be informed that the gold MR should contain a superlative operation (\ie, {\tt argmax}).
This motivates us to explore alternative forms of extra supervision other than fully-specified MRs.


Based on these three insights, we propose \method, a framework to merge active and weak supervision for semantic parsing.
\autoref{fig:wassp_overview} presents a schematic overview.
Specifically, \method iteratively performs the following steps:

\noindent
\textbf{Step 1: } Train the semantic parser $P(\hat{\mr}|\x;\theta)$ to convergence by optimizing Eq.~\ref{wssp} over the training set $D$ and discovered gold MR candidates $\hat{Z}$;

\noindent
\textbf{Step 2: } With active sample selection heuristic, select a small subset of the training set $D^{\prime}\subseteq D$ within certain budget and query the annotator for extra supervision for examples in $D^{\prime}$;

\noindent
\textbf{Step 3: } After receiving annotations, \eg, in the form of annotated MRs $\{ {\mr}_j^{*}| (\env_j, \x_j, \ans_j)\in D^{\prime}\}$, update the set of high-reward MRs $\hat{Z}$ explored so far for each training instance in $D^\prime$.
If the form of extra supervision is annotated MRs, this simply amounts to set $\hat{Z}$ to only contain the annotated MR, \ie, $\hat{Z} = \{ {\mr}^{*} \}$.
We will introduce other form of extra supervision in later sections.


The motivation behind \method is very simple; every time weakly-supervised training stalls, \method allows the parser to select a small fraction of the training set and query for extra supervision,
and use the received extra supervision to help continue the training process.
In the following sections, we examine two design decisions: (1) how to select our subset of data to query $D^{\prime}$ , and (2) what varieties of supervision can \method utilize.


%% file: sub-parts/3-sample-selection.tex
In this section, we examine strategies to select examples that are more likely to improve the performance of the semantic parser. First we propose a very simple yet effective \textit{correctness-based} method, then we investigate two types of commonly adopted heuristics in active learning, \textit{uncertainty-based} methods and \textit{coverage-based} methods.
 
\subsection{Correctness-based Method}
In traditional \textit{pool-based active learning} \citep{lewis1994sequential}, the learner selects from a pool of examples without any information about the true label of example.
However, for weakly-supervised semantic parsing we do have the expected execution result $\ans$ as indirect information that can be used to inform our choice of data to annotate.

As noted above, an MR executing to the correct result does not necessarily entail that the final logical form is correct ($\env(\hat{\mr})=\ans \not\Rightarrow \hat{\mr}=\mr$) because the MR may be spurious.
However, if an MR does not execute to the correct result, this \emph{does} entail that the MR is incorrect ($\env(\hat{\mr}) \not= \ans \Rightarrow \hat{\mr} \not= \mr$).
Taking advantage of this fact, we can derive a \textit{correctness-based heuristic}, which prefers to select the examples $\x$ for which there is no MR generated by the parser that matches the expected execution result thus no learning signal (\ie reward) is provided for the semantic parser.
These examples are much more likely to benefit from extra supervision than examples where the parser already has found at least one semantic parse corresponding to a correct answer. 

\subsection{Uncertainty-based Methods}
\label{section:uncertainty}
Uncertainty-based methods are a standard set of active learning techniques \citep{lewis1994heterogeneous,tong2001support,shen2004multi} that attempt to
maximize the \textit{informativeness} of selected examples by preferring examples where the learner is most uncertain about its prediction.
In the context of semantic parsing, we follow \citep{duong2018active} and measure the uncertainty on an example \x by the \textbf{least confidence score} \citep{culotta2005reducing}:
\begin{equation}
\label{eq:confidence}
    \x^{\prime} = \arg\min_{\x \in D} 
        [\max_{\Tilde{\mr}\in Z} P(\Tilde{\mr}|\x; \theta)]
\end{equation}

Though we can not enumerate all \mr in the MR space $Z$, it is easy to approximate this measurement by only enumerating all MRs remain in the beam $Z_B$ after beam search.

\subsection{Coverage-based Methods}
\label{section:coverage}
Coverage-based methods are another common family of methods in active learning \citep{dasgupta2008hierarchical}. This type of methods consider the \textit{representativeness} of the selected examples and attempts to cover as many other unselected examples as possible. 
Here we explore two different kinds of methods that adopt the idea of \textit{representativeness}:

\noindent
\textbf{Failed Word Coverage: }
Some particular words may be particularly difficult for semantic parsers to recognize, either because they are infrequent, or because they are only loosely connected to the MR. 
Based on this fact, we introduce a heuristic to select the examples with the largest number of words that are more prone to cause failure. 
An example is labeled as \texttt{fail} if the MR given by the semantic parser can not execute to the expected result and we denote the set of \texttt{fail} examples in the training set as $D^{-}\subseteq D$. Then given a bag-of-words representation for example $\x=[q^1,...,q^m]$ where $m$ is the vocabulary size, we estimate the probability of a word $q^j$ that leads to \texttt{fail} by counting the occurrences of this word in the \texttt{fail} examples and all examples:
\begin{equation*}
    P(\texttt{fail}|q^j) = \sum_{\x\in D^{-}} q^j /
    \sum_{\x\in D} q^j
\end{equation*}
Then we select the examples that covers more of words that are more likely to cause failure as following:
\begin{equation*}
\label{failed-words}
    \x^{\prime} = \arg\max_{\x \in D} \prod_{j=1}^{m} q^j \cdot P(\texttt{fail}|q^j)
\end{equation*}

\noindent
\textbf{Clustering: } 
For this method, we attempt to cluster together similar NL utterances.
For each example, a sentence-level embedding is computed by averaging the pre-trained GloVe embedding \citep{pennington2014glove} of the words, then we adopt the K-Means algorithm to perform clustering of the training examples.
Given a clustering of the training examples, we first rank the clusters by their size and leave out the last 20\% of the clusters to reduce the risk of selecting examples that are not representative (\ie outliers). Then from each of the remaining clusters, we random sample equal number of examples to encourage \emph{diversity}.

%% file: sub-parts/4-extra-supervision.tex
\begin{table}[tb]
  \centering
  \small
  \setlength\tabcolsep{2pt}
  \begin{tabular}{lp{5.8cm}}
  \toprule
  
\multicolumn{2}{l}{\textbf{NL Utterance \x:}~~{\it Which country is the \#1 gold medal winner?} } \\
  {\bf Full MR $\mr_f$:} &
\begin{lstlisting}[style=nsmcode]
(argmax all_rows `Gold') 
(hop v0 `Country') 
\end{lstlisting} \\
  {\bf MR Sketch $\mr_s$:} &
\begin{lstlisting}[style=nsmcode]
(argmax ...) (hop ...) 
\end{lstlisting} \\
  \bottomrule
  \end{tabular}
  \caption{An example of fully-specified MR and MR sketch}
  \label{tab:full-sketch}
\end{table}

In addition to the trivial solution of using fully-specified MRs as extra supervision, here we introduce an example of other forms of annotations that \method can utilize by merging them with the weakly-supervised training: the MR sketch. An example of a fully-specified MR and its sketch are shown in \cref{tab:full-sketch}

\noindent \textbf{Fully-specified MRs} Using fully-specified MRs as extra supervision for the selected query examples is an obvious strategy. We define the fully-specified MRs as a sequence of complete operations or function calls with all the variables or arguments and ready to be executed. An example of this is shown in \cref{tab:full-sketch} as $\mr_f$. A fully-specified MR can simply mount the explored high-reward MRs and be directly it used for training.

\noindent \textbf{MR Sketches} A \textit{sketch} of an executable MR is the sequence of operators or function names that does not fill in variables or arguments. An example MR sketch is shown in \cref{tab:full-sketch} as $\mr_s$. With MR sketch annotation, we can: 1) Remove spurious high-reward MRs where the sketch does not match the gold MR sketch from $\hat{Z}$; 2) Use this sketch as a guide for future exploration (\eg constrained decoding), the high-reward MRs are only saved in $\hat{Z}$ if their MR sketches match the gold MR sketch.
These sketches provide several benefits:

\noindent
\textbf{1) Reduction of search space.} Though MR sketches do not completely remove ambiguity, with the sketch, the parser only needs to fill out variables, and thus the search space is also greatly reduced. Moreover, since most neural semantic parsing models adopt a copy mechanism, variable type matching, or syntactic constraints \citep{krishnamurthy2017neural,liang2016neural}, the search space becomes even more restricted and leaves little room for spurious MRs making it much easier to explore. For example, all spurious forms are pruned when given the gold MR sketch of {\tt (argmax ...)(hop ...)} in \cref{fig:wassp_overview};

\noindent
\gn{Hmm, I don't know if it's very convincing that the annotator doesn't need to know the details of the grammar to annotate the sketch. At the very least, it'd be good to have an example of why this might be the case.}
\textbf{2) Increased generality over the MR.} 
Compared to annotating with fully-specified MRs (as $z_f$ in \cref{tab:full-sketch}),
a sketch is a higher-level abstraction, which resembles procedural knowledge for the model to parse this type of NL utterances without being deeply coupled with the specific details of the example itself. Take the example in \autoref{tab:full-sketch}, after annotating this example with its MR sketch {\tt (argmax ...)(hop ...)}, annotators can easily annotate other similar NL queries as ``Which country has the largest population'' or ``Which team was ranked the last'' without wasting time filling the detailed arguments. This special trait may speed up the manual annotation process thus could potentially be cheaper to obtain.


%% file: sub-parts/5-experiments.tex
\subsection{Experiment Setup}
\noindent \textbf{Dataset:}
We evaluate the performance of \method on two different datasets: WikiSQL \citep{zhong2017seq2sql} and WikiTableQuestions \citep{pasupat2015compositional}. The statistics of these two datasets are shown in \autoref{tab:dataset-stats}. 
The WikiSQL dataset provides both gold MRs and their expected execution results, making it possible to perform simulated experiments where we train the same model with weak, active, or full supervision. WikiTableQuestions does not have annotated MRs, and thus we perform only limited active learning experiments querying real human annotators to add additional MRs.


\begin{table}[tb]
    \centering
    \small
    \begin{tabular}{lcc}
        \hline
        Dataset             & WikiSQL   & WikiTableQuestions    \\
        \hline\hline
        \#Tables            & 26,531    & 2,108                 \\
        \#Questions         & 80,654    & 22,033                \\
        \hline
        NL Question         & \cmark    & \cmark                \\
        Table Content       & \cmark    & \cmark                \\
        Gold MR             & \cmark    & \xmark                \\
        Gold Exec. Result   & \cmark   & \cmark     \\
        \hline
    \end{tabular}
    \caption{Statistics of the datasets used in the experiments.}
    \label{tab:dataset-stats}
\end{table}

\noindent \textbf{Neural Semantic Parsing Model:} 
\label{section:nsm}
For the underlying neural semantic parsing model for \method, we adopt neural symbolic machines (NSM) \citep{liang2016neural}, a strong weakly-supervised semantic parser.
NSM uses a sequence-to-sequence network to transduce an input utterance into an MR. 
The recurrent decoder is augmented with a memory component to cache intermediate execution results in a partially generated MR (\eg, the result of {\tt argmax} for $\mr_0$ in~\autoref{fig:wassp_overview} is cached as $v_0$), which can be referenced by following operations (\eg, in the {\tt hop} function in $\mr_0$).
We follow~\citep{liang2018memory}, and train NSM using memory-augmented policy optimization (MAPO), which uses a memory buffer to cache currently explored high-reward MRs.

The NSM trained with MAPO is the state-of-the-art model for weakly-supervised semantic parsing on the WikiSQL dataset (detailed below); a single model (\ie without ensemble) can reach an execution accuracy of 72.4\% and 72.6\% on the dev and test set, respectively. 
Upon inspection, we found that the syntax rules of NSM did not have full coverage of the SQL queries in the WikiSQL dataset, and we further augmented the coverage of the NSM syntax rules, improving this to 75.2\% dev accuracy and 75.6\% test accuracy with pure weak supervision. On the WikiTableQuestions dataset, an ensemble of 10 such models achieves the state-of-the-art performance of 46.3\% execution accuracy while the performance of a single model is 43.1\%($\pm0.5\%$). 
We use this as a strong baseline which we aim to improve.

\noindent \textbf{Training Procedure:}
First we follow the procedure of \citet{liang2018memory} to train NSM with MAPO on both WikiSQL and WikiTableQuestions datasets with the same set of hyperparameters as used in the original paper. Then for WikiSQL, we run \method for 3 iterations with query budget 1,000 or more and only run for one iteration with smaller budget. For each iteration, the model queries for extra supervision and then it is trained for another 5K steps. The query budget is evenly distributed to these 3 iterations and limited by the total amount. For WikiTableQuestions, we simply run one such iteration (due to limit number of annotations obtained) but let it train for 50K steps with human annotated MRs. 

\noindent \textbf{Evaluation Metric:}
As with previous works \citep{pasupat2015compositional,zhong2017seq2sql,dong2018coarse,liang2018memory}, we measure the execution \emph{accuracy}, defined as the fraction of examples with correct execution results.

\subsection{Main Results}
\label{section:main-results}

\begin{table}[tb]
    \centering
    \small
    \begin{tabular}{lc}
        \hline
        Query Budget & Acc. (Imp.)\\\hline\hline
        0 (Pure Weak Supervision) & 75.6       \\\hdashline
        100 (0.2\%)         & 76.3(+0.7) \\
        200 (0.4\%)         & 76.7(+1.1) \\
        500 (0.9\%)         & 77.7(+2.1) \\
        1,000 (1.8\%)      & 78.6(+3.0) \\
        2,500 (4.4\%)      & 79.0(+3.4) \\
        10,000 (17.7\%)     & 79.8(+4.2) \\\hdashline
        56,355(All) & 80.3       \\\hline\hline
        \textbf{Previous Weak Supervision Methods} & Acc. \\
        NSM+MML \citep{liang2016neural} & 70.7 \\
        NSM+MAPO \citep{liang2016neural,liang2018memory} & 72.6  \\\hline
        \textbf{Previous Full Supervision Methods} & Acc. \\
        STAMP \citep{sun2018semantic} & 74.4 \\
        TranX+AP \citep{yin2018tranx} & 78.6 \\
        Coarse2Fine \citep{dong2018coarse} & 78.5 \\
        TypeSQL+TC \citep{yu2018typesql} & 82.6 \\\hline
    \end{tabular}
    \caption{\method with varying budget and previous fully- or weakly-supervised methods on WikiSQL. All methods use table contents during training and are evaluated on the test set.}
    \label{tab:acc-budget:wikisql}
\end{table}

\begin{table}[tb]
    \centering
    \small
    \begin{tabular}{lcc}
        \hline
        Query Budget & Cold Start & Warm Start\\\hline\hline
        0 (Pure Weak Supervision) & 8.5 & 42.7      \\\hdashline
        50 (0.5\%)         & 34.1(+25.6) & 42.7(+0.0) \\
        100 (0.9\%)      & 37.7(+29.2) & 43.2(+0.5) \\\hline\hline
    \end{tabular}
    \begin{tabular}{p{6.2cm}c}
        \textbf{Previous Weak Supervision Methods} & Acc. \\
        \citet{pasupat2015compositional} & 37.1 \\
        \citet{neelakantan2016learning} & 34.2 \\
        \citet{haug2018neural} & 34.8 \\
        \citet{zhang2017macro} & 43.7 \\\hline
    \end{tabular}
    \caption{\method with varying budget and different types of weakly-supervised training start on WikiTableQuestions. Evaluated by execution accuracy on the test set, improvements are noted in brackets.
    }
    \label{tab:acc-budget:wtq}
\end{table}

\noindent \textbf{Effectiveness of \method.}
First, to answer RQ1, 
we primarily investigate the performance of \method w.r.t.~the amount of extra supervision during training.
\autoref{tab:acc-budget:wikisql} and \autoref{tab:acc-budget:wtq} list results of \method with varying query budgets on WikiSQL and WikiTableQuestions respectively. On the WikiTableQuestions dataset, MAPO has the option to begin with an empty memory buffer (\ie a cold start) or warm up the memory buffer by searching with manually-designed pruning rules as heuristics (\ie a warm start). Since designing the pruning heuristics also includes non-trivial human effort, we include the both results from a model trained from a cold start and a warm start to have a rather complete study. 
Our model presented here uses the correctness-based query selection heuristic and fully-specified MRs as extra supervision.
We also compare \method with existing weakly- or fully-supervised methods. 

From \autoref{tab:acc-budget:wikisql} we can see that with annotating only 100 (0.2\%) examples from the training set, \method can improve the performance by 0.7\% while querying 500 examples (0.89\%) achieves an absolute improvement of 2.1\% over the same model trained with pure weak supervision. 
Notably, if we allow a query budget of 1,000 examples, \method achieves an execution accuracy of 78.6\%, a 6.0\% absolute improvement over previous state-of-the-art result (72.6\%) with only weak supervision \citep{liang2018memory}.
Additionally, with only $1.8\%$ of annotated training examples, \method is already quite competitive against some of the strong fully-supervised systems trained with the whole annotated training set ~\citep{sun2018semantic,yin2018tranx,dong2018coarse}. 
This result is also only 1.7\% away from the same semantic parser (NSM) trained with full supervision (80.3\%).
Finally, if we further increase the query budget to 10,000 examples, \method could achieve similar performance comparable with training the semantic parser using full supervision, but with 80\% less annotated data. This set of results shows that \method allows learning with significantly less supervision. 

We also have some interesting findings for the WikiTableQuestions dataset. While \autoref{tab:acc-budget:wtq} shows that \method can improve the performance of a model trained with either cold or warm start, \method yields a substantial 
on models trained with a cold start: a 25.6\% improvement with only 50 examples. Further investigations shows that the baseline model trained with a cold start on WikiTableQuestions is only able to correctly parse simple NL queries (\eg count the number of rows, find maximum number, etc) that can be addressed with one statement (\eg {\tt (count ...), {\tt (maximum ...)}}). Since the model is under-trained, the probability of it exploring a high-reward complex MR for a sophisticated questions is very low. In this way, injecting a few annotations for the examples the model can not solve yet makes the model realize that more rewards may be obtained by more complex MRs, which in turn motivates and guides the exploration of the model. 



\noindent \textbf{Query Sample Selection Heuristics.}
To answer RQ2, we evaluate \method trained with different sample selection heuristics introduced previously
 and summarize the results in \autoref{tab:heuristic-acc}.
We found the simple correctness-based heuristic is effective, perhaps surprisingly so.
Using uncertainty scores alone does not achieve good performance, barely exceeding random selection. 
Further investigation finds that by the end of weakly-supervised training, 84.0\% of examples in training set with their parsed MRs not executed to the expected results have a confidence (Eq.~\ref{eq:confidence}) over 0.9.
This finding suggests that incorrect MRs could still have high model-assigned probability, which is worth future investigation.

Nevertheless, \method achieves the best performance by combining the two heuristics (\ie, selecting the most uncertain examples among the failed ones)\footnote{We also tried to combine other sets of heuristics but had equivalent or worse results than using one of the components.}. 
This query sample selection method combines the idea of correctness and uncertainty to maximizes the informativeness of the selected examples and reaches 78.4\% and 79.0\% of execution accuracy on dev. and test set.
Finally, we found the coverage-based methods do not work as well, which may result from our relatively simple approach of measuring coverage using sentence embedding or failed words.\footnote{A stronger sentence embedding method may improve the performance, but a discussion of the best sentence embedding method is out of the scope of this paper.}

\begin{table}[tb]
\begin{center}
\small
\begin{tabular}{lcc}
\hline
Selection Heuristics           & Dev.     & Test    \\ \hline\hline
Baseline (No query made)       & 75.3 & 75.6 \\\hdashline
Random                         & 76.2 & 76.4 \\
Correctness                    & 78.3 & 78.6 \\ 
Uncertainty                    & 76.4 & 77.0 \\ 
\;\;\;\; +Correctness          & \textbf{78.4} & \textbf{79.0} \\
Coverage-based (Fail Words)    & 77.1 & 77.4 \\ 
Coverage-based (Clustering)    & 77.8 & 77.5 \\\hline
\end{tabular}
\end{center}
\caption{Comparison of Different Query Sample Selection Heuristics. An equal query budget of 1,000 examples for gold MRs is given for all methods, performance measured by execution accuracy. Best numbers are in bold.}
\label{tab:heuristic-acc}
\end{table}

\noindent \textbf{Forms of Supervision}
Here we study how well \method can adapt to other forms of supervision besides the gold fully-specified MR. To answer RQ3, we conduct experiments with the MR sketch as extra supervision and compare its performance to using the fully-specified MR under the framework of \method and the results are shown in \autoref{tab:form-acc:wikisql} and \autoref{tab:form-acc:wtq} for the WikiSQL and WikiTableQuestions datasets, respectively. In this set of experiments, we annotate examples selected with \textit{correctness} heuristic.

From these tables, we can see that the performance of a model learnt from MR sketches as extra supervision is comparable to the model learnt from fully-specified MR, regardless of the budget is larger (1,000 to 10,000, maximum gap of 1.4\% on WikiSQL) or smaller (50 to 100, maximum gap of 2.4\% on WikiTableQuestions). 
This shows that fully-specified MRs are not the only option as extra supervision for \method to achieve good improvement over weakly-supervised semantic parsers. 
We hope in future work to investigate other forms of extra supervision that may benefit \method.


\begin{table}[tb]
\begin{center}
\small
\begin{tabular}{lcc}
\hline
Query Budget            & Full MR & MR Sketch\\ \hline\hline
1,000 (1.8\%)     & 78.6    &  77.2 (-1.4)    \\
2,500 (4.4\%)     & 79.0    &  78.4 (-0.6)   \\
10,000 (17.7\%)   & 79.8    &  78.9 (-0.9)   \\\hdashline
56,355 (All)            & 80.3    &  79.1 (-1.2)   \\\hline
\end{tabular}
\end{center}
\caption{Comparison of supervision forms for WikiSQL. Numbers in the brackets show the gap between full and sketched MRs.}
\label{tab:form-acc:wikisql}
\end{table}

\begin{table}[t]
\begin{center}
\small
\begin{tabular}{lcc}
\hline
Query Budget            & Full MR & MR Sketch\\ \hline\hline
50 (0.5\%)     & 34.1    &  33.2 (-0.9)    \\
100 (0.9\%)     & 37.7    &  35.3 (-2.4)   \\\hline
\end{tabular}
\end{center}
\caption{Comparison of Extra Supervision Forms for WikiTableQuestions. Numbers in the brackets show the gap between full and sketched MRs. Weakly-supervised training starts from a cold start.}
\label{tab:form-acc:wtq}
\end{table}

\input{sub-parts/case-study.tex}

%% file: sub-parts/case-study.tex
\begin{table}[tb]
  \centering
  \small
  \setlength\tabcolsep{2pt}
  \begin{tabular}{lp{7cm}}
  \toprule
  
\multicolumn{2}{l}{\textsc{Selected-1}~~{\it \underline{How many visitors} in 2007 \underline{were there}?} } \\
  {$\mr_f$} &
\begin{lstlisting}[style=nsmcode]
(filter_eq all_rows `2007' `Year') 
(count v0) 
\end{lstlisting} \\
  {$\mr_s$} &
\begin{lstlisting}[style=nsmcode]
(filter_eq ...) (count ...) 
\end{lstlisting} \\\hdashline

\multicolumn{2}{l}{\textsc{Tested-1}~~{\it \underline{How many} CIL competitions \underline{were there}?} } \\
  {$\hat{\mr}$} &
\begin{lstlisting}[style=nsmcode]
(filter_eq all_rows `CIL' `Competition') 
(hop v0 `established') |\xmark|
\end{lstlisting} \\
  {$\hat{\mr}^{\prime}$} &
\begin{lstlisting}[style=nsmcode]
(filter_eq all_rows `CIL' `Competition') 
(count v0) |\cmark|
\end{lstlisting} \\\hline

  \multicolumn{2}{l}{\textsc{Selected-2}~~{\it How many points did the \underline{driver who won}} } \\
  & {\it~~~~~~~~~~ \underline{\$127,541} driving car \#31 get?} \\
  {$\mr_f$} &
\begin{lstlisting}[style=nsmcode]
(filter_eq all_rows `$127,541' `Winning') 
(filter_eq v0 `#31' `Car') (hop v1 `Pts') 
\end{lstlisting} \\
  {$\mr_s$} &
\begin{lstlisting}[style=nsmcode]
(filter_eq ...)(filter_eq ...)(hop ...) 
\end{lstlisting} \\\hdashline

  \multicolumn{2}{l}{\textsc{Tested-2}~~{\it Which \underline{driver won \$40,000} in the NW Cup?} } \\
  {$\hat{\mr}$} &
\begin{lstlisting}[style=nsmcode]
(filter_eq all_rows `NW Cup' `Series') 
(hop v0 `driver') |\xmark|
\end{lstlisting} \\
  {$\hat{\mr}^{\prime}$} &
\begin{lstlisting}[style=nsmcode]
(filter_eq all_rows `$40,000' `Winning')
(filter_eq all_rows `NW Cup' `Series')
(hop v0 `driver') |\cmark|
\end{lstlisting} \\\hline

  \multicolumn{2}{l}{\textsc{Selected-3}~~{\it What position was WIC \underline{in a year}} } \\
  & {\it~~~~~~~~~~ \underline{later than 2008}?} \\
  {$\mr_f$} &
\begin{lstlisting}[style=nsmcode]
(filter_eq all_rows `WIC' `Competition') 
(filter_greater v0 `2008' `Year') 
(hop v1 `Position') 
\end{lstlisting} \\
  {$\mr_s$} &
\begin{lstlisting}[style=nsmcode]
(filter_eq ...)(filter_greater ...)
(hop ...) 
\end{lstlisting} \\\hdashline

  \multicolumn{2}{l}{\textsc{Tested-3}~~{\it What Chassis has \underline{a year later than 1989}?} } \\
  {$\hat{\mr}$} &
\begin{lstlisting}[style=nsmcode]
(filter_less all_rows `1989' `Year') 
(hop v0 `Chassis') |\xmark|
\end{lstlisting} \\
  {$\hat{\mr}^{\prime}$} &
\begin{lstlisting}[style=nsmcode]
(filter_greater all_rows `1989' `Year') 
(hop v0 `Chassis') |\cmark|
\end{lstlisting} \\
  \bottomrule
  \end{tabular}
  \caption{Examples selected to query (\textsc{Selected-}) and similar test examples (\textsc{Tested-}) with their top-1 parsed MR before ($\hat{\mr}$) and after ($\hat{\mr}^{\prime}$) \method is applied. $\mr_f$ and $\mr_s$ denote full and sketched MR as extra supervision. Similar parts between query examples and test examples are underlined.}
  \label{tab:case-study}
  \vspace{-5mm}
\end{table}

\subsection{Qualitative Analysis}
Here we conduct a qualitative analysis on how the examples selected and queried by \method help it generalize to other unseen or not queried examples. Example 1 from \autoref{tab:case-study} shows that extra supervision for the query set can generalize to other examples with similar NL utterance structure. For Example 2, the missing filter is corrected after receiving extra supervision for an example with similar conditions. As in Example 3, the parser can not use the correct filter for comparing time, but as the example above being selected and queried, the semantic parser trained with extra supervision learns to map ``later than" to {\tt filter\_greater} for comparing time. These examples show that extra supervision for the actively queried examples, helps the model with \method learn to generalize to other similar test examples.

%% file: sub-parts/related-work.tex
\noindent
\citep{artzi2013weakly} proposed to jointly learn from meaning and context for. In the work of \citep{krishnamurthy2012weakly}, they combined weak supervision from the knowledge base and the NL sentences. To handle the problem of spurious forms, \citep{muhlgay2019value} proposed a value-based search method with a trained critic network trained with the environment. 
In \citep{shen2004multi}, they explored different active learning heuristics for named entity recognition, including informativeness, representativeness and diversity. 
\citep{duong2018active} studied active learning for fully-supervised semantic parsing and showed that an uncertainty-based active learning method is powerful for traditional data collection but not useful to overnight data collection.